\documentclass[sigconf]{acmart}
\AtBeginDocument{%
  }

\usepackage{booktabs}
\usepackage{xcolor}
\usepackage{orcidlink}

\AtBeginDocument{%
  }

\copyrightyear{2026}
\acmYear{2026}
\setcopyright{cc}
\setcctype{by}
\acmConference[SA Technical Communications '26]{SIGGRAPH Asia 2026 Technical Communications}{December 01--04, 2026}{Kuala Lumpur, Malaysia}
\acmBooktitle{SIGGRAPH Asia 2026 Technical Communications (SA Technical Communications '26), December 01--04, 2026, Kuala Lumpur, Malaysia}
\acmDOI{10.1145/3829339.3847862}
\acmISBN{979-8-4007-2841-9/2026/12}

\acmSubmissionID{tcom\_205}

\begin{document}

\title[Scene-to-Vector Sketching via Decoupled Structure and Texture Optimization]{SketchForge: Training-Free Scene-to-Vector Sketching via Decoupled Structure and Texture Optimization}

\author{Quang-Thinh Nguyen}
\authornote{Both authors contributed equally to this research.}
% \email{22120346@student.hcmus.edu.vn}
\orcid{0009-0002-1283-8663}
\author{Duy-Khang Do}
\authornotemark[1]
% \email{22120454@student.hcmus.edu.vn}
\orcid{0009-0003-7430-1586}
\author{Hien Trinh}
% \email{23120254@student.hcmus.edu.vn}
\orcid{0009-0006-0396-7219}
\affiliation{%
  \institution{University of Science}
  \city{Ho Chi Minh}
  \country{Vietnam}
}
\affiliation{%
  \institution{Vietnam National University}
  \city{Ho Chi Minh}
  \country{Vietnam}
}

\author{Tam V. Nguyen}
\orcid{0000-0003-0236-7992}
\affiliation{%
  \institution{University of Dayton}
  \city{Dayton}
  \state{Ohio}
  \country{US}
}
% \email{tamnguyen@udayton.edu}

\author{Minh-Triet Tran}
% \email{tmtriet@fit.hcmus.edu.vn}
\orcid{0000-0003-3046-3041}
\author{Trung-Nghia Le}
\email{ltnghia@fit.hcmus.edu.vn}
\orcid{0000-0002-7363-2610}
\correspondingauthor
\affiliation{%
  \institution{University of Science}
  \city{Ho Chi Minh}
  \country{Vietnam}
}
\affiliation{%
  \institution{Vietnam National University}
  \city{Ho Chi Minh}
  \country{Vietnam}
}

%%
%% By default, the full list of authors will be used in the page
%% headers. Often, this list is too long, and will overlap
%% other information printed in the page headers. This command allows
%% the author to define a more concise list
%% of authors' names for this purpose.
\renewcommand{\shortauthors}{Quang-Thinh Nguyen et al.}

\begin{abstract}
Converting a raster image into a resolution-independent, stroke-editable vector sketch while faithfully preserving both structural contours and tonal texture remains challenging: raster-space rendering pipelines are difficult to edit at the stroke level, while generation-from-scratch vectorization methods struggle to remain faithful to an arbitrary input image. We present SketchForge, a training-free framework that converts an input raster image into a resolution-independent SVG sketch through per-instance test-time optimization of Bézier curves, requiring no additional training or fine-tuning. Strokes are seeded via a contour and attention-guided initialization, then refined by a vector-native differentiable rendering pipeline that decouples structural contour strokes from free-form texture strokes under dedicated objectives, composited via multiply-blend layering. Extensive experiments show that SketchForge consistently outperforms existing vector-based baselines in structural fidelity and generation quality. Project page: \url{https://khangcntn123.github.io/SketchForge}
\end{abstract}




\begin{teaserfigure}
  \centering
    \includegraphics[width=\linewidth]{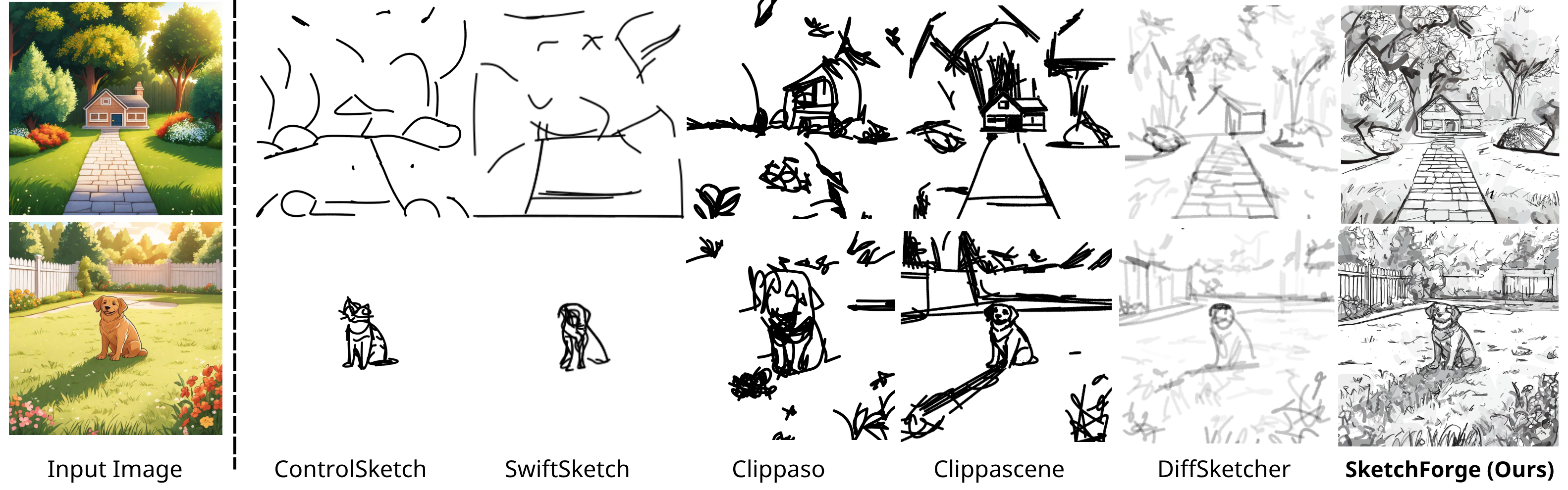}
    \caption{\textbf{Qualitative comparison on sketch vectorization.} ControlSketch~\cite{arar2025swiftsketch} and SwiftSketch~\cite{arar2025swiftsketch} produce sparse, abstract strokes; CLIPasso~\cite{vinker2022clipasso} yields spatially disordered strokes; CLIPascene~\cite{10378256} over-abstracts fine detail; adapted DiffSketcher~\cite{xing2023diffsketcher} under-represents local texture. Our SketchForge most faithfully preserves object boundaries and structural detail.}
    \Description{A grid comparing six vector-based sketch methods across
    two content images. The first content image shows a park walkway
    scene with trees; the second shows a dog resting on a lawn beside a
    wooden fence. Each subsequent column shows one method's output for
    both scenes. ControlSketch and SwiftSketch: sparse disconnected
    lines. CLIPasso: denser but disordered strokes. CLIPascene: bold
    silhouettes with lost detail. DiffSketcher: lighter, softer strokes.
    SketchForge: the most detailed, structurally faithful outputs.}
  \label{fig:teaser}
\end{teaserfigure}

\maketitle

\section{Introduction}

Sketching conveys structure and style through sparse strokes, yet transferring stroke-level appearance to raster content without losing geometry remains difficult. Stroke-based methods~\cite{vinker2022clipasso,xing2023diffsketcher} often generate from scratch, while raster diffusion pipelines~\cite{rombach2022high,zhang2023adding,ye2023ip} do not provide editable vector strokes.

DiffVG~\cite{DBLP:journals/tog/LiLGR20} introduced differentiable vector rasterization, enabling end-to-end stroke optimization and CLIP-supervised sketch synthesis~\cite{DBLP:conf/nips/FransSW22,vinker2022clipasso,10378256}. Subsequent work has explored text-to-SVG generation with diffusion priors, including VectorFusion~\cite{jain2023vectorfusion} and SVGDreamer~\cite{10658565}; autoregressive and latent-sequence generation~\cite{11094922,DBLP:conf/aaai/ZhouZYXH26}; and improvements through stroke pruning~\cite{DBLP:conf/aaai/HirschornJA24}, differentiable splatting~\cite{liu2025bzier}, semantic region guidance~\cite{DBLP:conf/eccv/ZhouZW24}, discrete token prediction~\cite{DBLP:journals/corr/abs-2504-06263}, and direct raster-to-sketch prediction~\cite{arar2025swiftsketch}. Vector customization further supports exemplar-preserving editing and style-controlled generation~\cite{10.1145/3610548.3618232,10.1145/3721238.3730707}, while AutoSketch~\cite{10.1145/3757377.3763828} completes partial sketches using automatically derived style descriptions. Despite this progress, faithfully translating a complete raster illustration into an editable vector sketch remains underexplored. In particular, optimizing all strokes toward a unified objective entangles two competing goals: contours require precise geometric alignment, whereas texture requires flexible strokes that capture tonal appearance.

% DiffVG~\cite{DBLP:journals/tog/LiLGR20} enabled end-to-end stroke optimization and CLIP-supervised sketching~\cite{DBLP:conf/nips/FransSW22,vinker2022clipasso,10378256}. Subsequent work introduced text-to-SVG diffusion~\cite{jain2023vectorfusion,10658565}, autoregressive or latent-sequence generation~\cite{11094922,DBLP:conf/aaai/ZhouZYXH26}, stroke reduction~\cite{DBLP:conf/aaai/HirschornJA24}, faster rendering~\cite{liu2025bzier}, region guidance~\cite{DBLP:conf/eccv/ZhouZW24}, discrete SVG tokens~\cite{DBLP:journals/corr/abs-2504-06263}, and direct image-to-vector prediction~\cite{arar2025swiftsketch}. Unlike these generation or completion settings, including vector customization~\cite{10.1145/3610548.3618232,10.1145/3721238.3730707} and AutoSketch~\cite{10.1145/3757377.3763828}, our goal is faithful conversion of a complete raster image without text or style descriptions.

\begin{figure*}[t!]
    \centering
    \includegraphics[width=\textwidth]{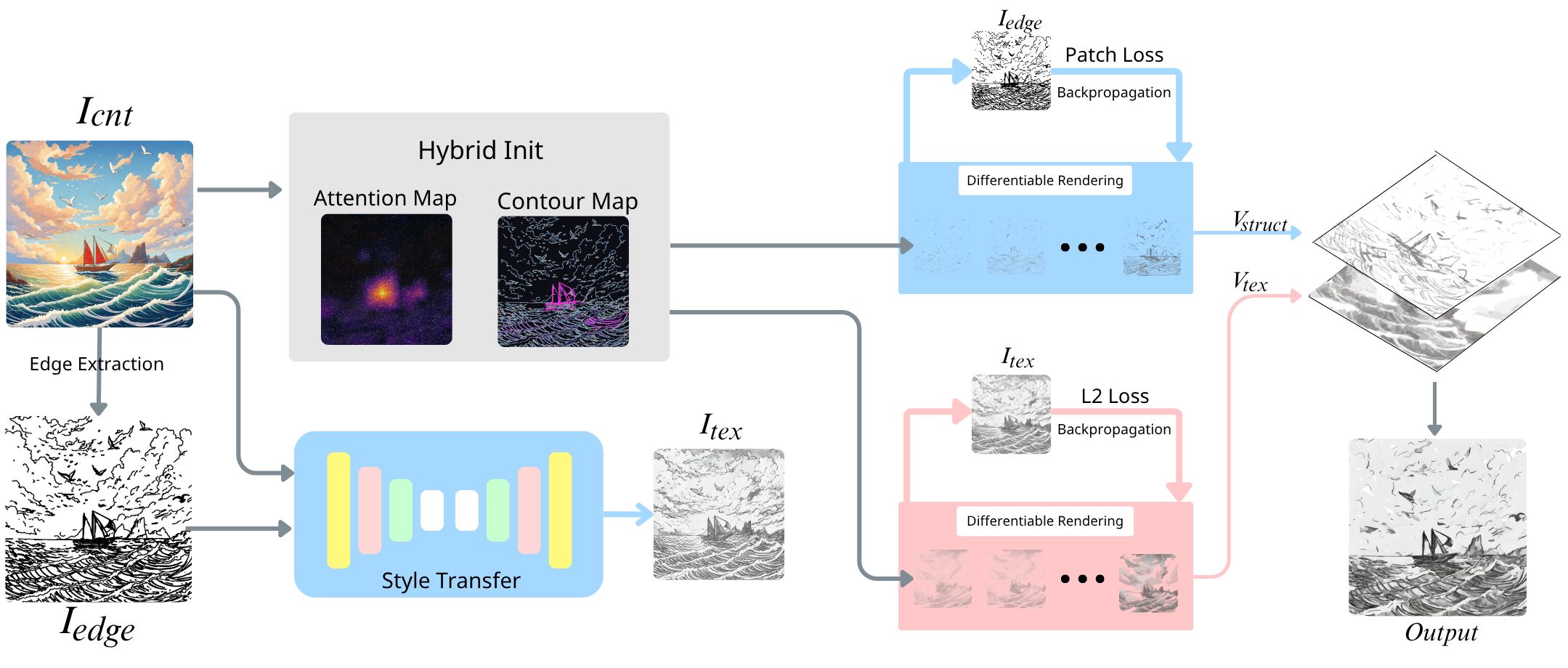}
    \vspace{-5mm}
    \caption{\textbf{SketchForge pipeline.} Guided initialization seeds two stroke sets: the structure branch optimizes geometry against $I_{\text{edge}}$ with a patch loss, while the texture branch optimizes all stroke attributes against $I_{\text{tex}}$ with an $L_2$ loss. Multiply blending produces the final SVG.}
    \Description{A pipeline diagram flowing left to right. On the far left, a
    colorful painted sailboat scene on turbulent waves under a golden sunset
    sky, labelled I\_cnt, sends one arrow rightward into a grey-bordered box
    labelled Hybrid Init, and another arrow downward through an Edge
    Extraction step into a black-and-white line-art rendition of the same
    scene, labelled I\_edge, showing sketched birds, sailboat, and wave
    contours. The Hybrid Init box contains two thumbnails: a dark Attention
    Map with a bright orange-purple heat-map glow near the sun, and a dark
    Contour Map with the sailboat and wave outlines highlighted in magenta.
    Below Hybrid Init, both I\_cnt and I\_edge feed into a blue Style
    Transfer block containing a sequence of colored vertical bar icons,
    which outputs a grayscale pencil-sketch-like rendition of the sailboat
    scene labelled I\_tex. From the initialization module, the strokes
    branch into two parallel optimization rows on the right. The upper,
    blue-tinted row shows I\_edge above a Patch Loss and Backpropagation
    label, feeding into a blue Differentiable Rendering block containing a
    sequence of small grayscale stroke-render thumbnails connected by an
    ellipsis, producing V\_struct. The lower, pink-tinted row shows I\_tex
    above an L2 Loss and Backpropagation label, feeding into a pink
    Differentiable Rendering block with its own sequence of small
    stroke-render thumbnails, producing V\_tex. On the far right, V\_struct
    and V\_tex are shown as two overlapping diamond-shaped translucent
    layers, which merge via a downward arrow into the final Output: a
    detailed grayscale SVG sketch of the sailboat scene with visible bird,
    wave, and rigging strokes.}
    \label{fig:loop}
\end{figure*}

We present \textbf{SketchForge}, a training-free framework that converts a full-color raster illustration into a resolution-independent SVG through per-instance optimization of Bézier curves (Fig.~\ref{fig:teaser}). SketchForge requires neither task-specific fine-tuning nor a pretrained diffusion backbone. It first extracts an edge map as the structural target and initializes strokes using complementary geometric and semantic cues. Contour-guided warping aligns strokes with prominent boundaries, while DINO attention~\cite{9709990} directs additional strokes toward salient regions. A differentiable optimization pipeline built on DiffVG~\cite{DBLP:journals/tog/LiLGR20} then separates the strokes into structural and textural layers. The structural layer is optimized with a patch-based multi-scale contour objective, whereas the textural layer reconstructs the global appearance of the input. Multiply-blend compositing integrates both layers into the final SVG. Experiments demonstrate that our proposed method outperforms existing vector-based methods in structural fidelity and perceptual quality. 
%We also showcase a set of applications, demonstrating the flexibility of the resulting vector representation of our method (Sec.~\ref{sec:applications}).

Our contributions are as follows:
\begin{itemize}
    \item We introduce a {training-free raster-to-vector sketching framework} that produces editable, resolution-independent SVGs through per-instance Bézier-curve optimization.
    \item We propose a {decoupled structure--texture formulation} that combines a patch-based multi-scale contour objective with global appearance reconstruction through multiply-blend compositing.
    \item We present a {contour and attention-guided initialization strategy} that places strokes at geometrically and semantically informative locations, improving optimization over uniform random initialization.
\end{itemize}

\section{Proposed Method}
\label{sec:method}

\subsection{Overview}
\label{sec:overview}

Given a full-color content image $I_{\text{cnt}}$, SketchForge proceeds in four stages (Fig.~\ref{fig:loop}). First, it extracts an edge target $I_{\text{edge}}$. Second, style transfer combines $I_{\text{cnt}}$ with $I_{\text{edge}}$ to produce the texture target $I_{\text{tex}}$. Third, guided initialization seeds independent stroke sets, optimized into $\mathcal{V}_{\text{struct}}$ and $\mathcal{V}_{\text{tex}}$. Finally, the two sets are fused into an editable SVG.

\subsection{Pre-processing}
\label{sec:preprocess}

We extract a black-on-white $I_{\text{edge}}$ from $I_{\text{cnt}}$ and normalize its line width. This map guides stroke placement, supervises structure optimization.

\subsection{Style Transfer}
\label{sec:style}

Before vector optimization, a StyleID-inspired module~\cite{DBLP:conf/cvpr/ChungHH24} injects reference features into DDIM-inverted~\cite{song2020denoising} content representations, with AdaIN~\cite{8237429} aligning latent statistics. The resulting raster $I_{\text{tex}}$ transfers the tonal and textural character of $I_{\text{edge}}$ while retaining the layout of $I_{\text{cnt}}$. With normalized edges as the default reference, this stage produces a sketch-like target without training, fine-tuning, or additional user input.

\subsection{Contour and Attention-Guided Initialization}
\label{sec:init}

SketchForge initializes $N$ strokes from both geometric and semantic cues rather than uniform random locations. A fraction $\rho$ is contour-guided: two-thirds are placed directly on detected contours and one-third interpolate adjacent placements. The remaining strokes are attention-guided to cover salient regions not explained by strong edges.

\paragraph{Contour-guided placement.}
Contours are extracted from $I_{\text{edge}}$ by thresholding and connected components, then weighted by arc length. Each stroke is warped to local geometry by mapping control-point residuals $\mathbf{r}_j$ from the principal axis $\mathbf{u}_k$ to the contour normal $\mathbf{n}(t_j)$:
\begin{equation}
    \tilde{p}_j = \mathcal{C}_k(t_j) + \beta\,\mathrm{sgn}(\mathbf{r}_j \cdot \mathbf{n}(t_j))\,\|\mathbf{r}_j\|\,\mathbf{n}(t_j),
    \qquad \beta = 0.3.
    \label{eq:warp}
\end{equation}
Smoothstep interpolation fills gaps, while pairs with dissimilarity below $\tau=0.3$ are skipped.

\paragraph{Attention-guided placement.}
A DINO~\cite{9709990} class-token attention map $\mathbf{A}$ and edge-confidence map $\mathbf{E}$ are power-transformed ($\gamma_{\mathbf A}=0.5$, $\gamma_{\mathbf E}=0.7$), averaged, and normalized into:
\begin{equation}
    \mathbf{P} = \frac{1}{2}\bigl(\mathbf{A} + \mathbf{E}\bigr).
\end{equation}
Locations sampled from $\mathbf P$ seed template strokes with small random offsets. The final pool combines direct contour placements, interpolated strokes, and attention samples, providing broad coverage before differentiable optimization.

\subsection{Stroke Optimization}
\label{sec:optim}

\paragraph{Structure optimization.}
With width and color fixed, control points are optimized toward $I_{\text{edge}}$. Because edges are spatially sparse, we use a multi-scale loss over a $G\times G$ grid:
\begin{equation}
    \mathcal{L}_{\text{struct}} =
    \frac{1}{G^2} \sum_{i,j}
    \left[
        \mathcal{L}_{fine}(\hat{I}_{ij}, I^e_{ij})
        + \mathcal{L}_{coarse}(\downarrow_r \hat{I}_{ij}, \downarrow_r I^e_{ij})
    \right],
    \label{eq:struct_loss}
\end{equation}
where $G=4$ and $\downarrow_r$ is average pooling with $r=4$. The fine term supplies stable gradients around sharp edge transitions, while the coarse term anchors low-frequency stroke mass when a patch contains few foreground pixels. Their combination yields $\mathcal{V}_{\text{struct}}$ aligned with $I_{\text{edge}}$.

\paragraph{Texture optimization.}
A separately initialized set frees control points, widths, opacity, and colors and reconstructs $I_{\text{tex}}$ globally:
\begin{equation}
    \mathcal{L}_{\text{tex}} = \|\mathcal{R}(\{s_i\}) - I_{\text{tex}}\|_2^2,
    \label{eq:tex_loss}
\end{equation}
producing $\mathcal{V}_{\text{tex}}$. This full-canvas objective captures tonal variation without forcing texture into sparse edge patches. Both branches use Adam with separate parameter groups for control points, color/opacity, and width; the structure branch keeps appearance fixed, whereas the texture branch optimizes all attributes.

\subsection{Layer-wise Multiply Fusion}
\label{sec:fusion}

The final SVG stacks $\mathcal{V}_{\text{tex}}$ below $\mathcal{V}_{\text{struct}}$ and applies \texttt{mix-blend-mode: multiply} to the structure layer, so $c^*_{xy}=c^s_{xy}\odot c^t_{xy}$. Dark contours reinforce local structure, while white regions preserve texture.

\section{Experiments}

\subsection{Experimental Settings}

\paragraph{Implementation Details.}
We initialize 700 Bézier strokes per branch on a $1024\times1024$ canvas and optimize them for 500 Adam iterations. The initial stroke width is $1.5$ pixels (maximum $100$). Geometry, color, and width use learning rates of $1.0$, $0.01$, and $0.1$, respectively. The structure branch optimizes geometry only, while the texture branch jointly optimizes geometry, width, opacity, and color.

\paragraph{Dataset. }

We evaluated 64 diverse raster illustrations generated with Stable Diffusion~\cite{rombach2022high}, spanning common daily objects to organic subjects such as animals and human figures with varying visual complexity.

\begin{table}[t!]
\centering
\caption{\textbf{Sketch vectorization on 64 content images.} DINO and DreamSim use $I_{\text{cnt}}$; Edge F1 uses $I_{\text{edge}}$. Best in \textbf{bold}.}
\label{tab:cmp_scene}
\vspace{-2mm}
\resizebox{\linewidth}{!}{%
\begin{tabular}{l cccc}
\toprule
\textbf{Method}
  & \textbf{Aesthetic} $\uparrow$
  & \textbf{DINO} $\uparrow$
  & \textbf{DreamSim} $\downarrow$
  & \textbf{Edge F1} $\uparrow$ \\
\midrule
CLIPasso~\cite{vinker2022clipasso}
  & 4.1690 & 0.2064 & 0.7256 & 0.5262 \\
CLIPascene~\cite{10378256}
  & 4.3904 & 0.2022 & 0.7024 & 0.5797 \\
SwiftSketch~\cite{arar2025swiftsketch}
  & 3.9784 & 0.1566 & 0.8305 & 0.3139 \\
ControlSketch~\cite{arar2025swiftsketch}
  & 4.1493 & 0.1470 & 0.7960 & 0.3299 \\
DiffSketcher~\cite{xing2023diffsketcher} (adapted)
  & 4.8020 & 0.3760 & \textbf{0.5351} & 0.8325 \\
\textbf{SketchForge (Ours)}
  & \textbf{4.8407} & \textbf{0.4507} & 0.5552 & \textbf{0.9780} \\
\bottomrule
\end{tabular}%
}
\vspace{-3mm}
\end{table}

\paragraph{Evaluation Metrics.}

We evaluate methods on four aspects: \textbf{Perceptual quality} (\textit{e.g.}, Aesthetic Score~\cite{schuhmann2022improvedaesthetic}, LPIPS~\cite{8578166}, and DreamSim~\cite{fu2023dreamsim}), \textbf{Semantic fidelity} (\textit{e.g.}, DINO~\cite{9709990}), \textbf{Structural fidelity} (\textit{e.g.}, SSIM~\cite{1284395}, MAE, and Edge F1), and \textbf{Geometric precision} (\textit{e.g.}, Chamfer distance~\cite{Barrow1977ParametricCA} on point sets sampled from skeletonized stroke paths).

\subsection{Comparison against State-of-the-Art}

We compare against CLIPasso~\cite{vinker2022clipasso}, CLIPascene~\cite{10378256}, SwiftSketch and ControlSketch~\cite{arar2025swiftsketch}, and an adapted version of DiffSketcher~\cite{xing2023diffsketcher}, where text conditioning is replaced by the input content image $I_{\text{cnt}}$.

Fig.~\ref{fig:teaser} shows that prior methods produce sparse or disordered strokes (ControlSketch and SwiftSketch), over-abstract details (CLIPasso and CLIPascene), or under-represent local textures (DiffSketcher). SketchForge better preserves object boundaries and fine structural details. Quantitatively, SketchForge achieves the highest Aesthetic Score (4.8407), DINO similarity (0.4507), and Edge F1 (0.9780), while obtaining a competitive DreamSim distance of 0.5552, outperforming previous vectorization approaches on overall perceptual, semantic, and structural quality.

\subsection{Ablation Study}

\paragraph{Stroke initialization.}
Tab.~\ref{tab:abl_init_struct} and Tab.~\ref{tab:abl_init_tex} compare uniform random sampling with our guided initialization on the isolated structure and texture branches. Guided placement improves every metric: for structure, Edge F1 rises from 0.7731 to 0.8841 and SSIM from 0.1120 to 0.8068; for texture, it improves DINO from 0.6378 to 0.7328 while reducing DreamSim and LPIPS. This confirms that semantically grounded initialization benefits both optimization branches.

\begin{table}[t!]
\centering
\caption{\textbf{Structure initialization ablation} against $I_{\text{edge}}$. Best in \textbf{bold}.}
\label{tab:abl_init_struct}
\vspace{-2mm}
\resizebox{0.7\linewidth}{!}{
\begin{tabular}{l c c c c}
\toprule
\textbf{Init.}
  & \textbf{MAE} $\downarrow$
  & \textbf{SSIM} $\uparrow$
  & \textbf{Chamfer} $\downarrow$
  & \textbf{Edge F1} $\uparrow$ \\
\midrule
Random
  & 0.4384 & 0.1120 & 0.0006 & 0.7731 \\
\textbf{Guided}
  & \textbf{0.2041} & \textbf{0.8068} & \textbf{0.0003} & \textbf{0.8841} \\
\bottomrule
\end{tabular}
}
\end{table}

\begin{table}[t!]
\centering
\caption{\textbf{Texture initialization ablation} against $I_{\text{tex}}$; Edge F1 uses $I_{\text{edge}}$. Best in \textbf{bold}.}
\label{tab:abl_init_tex}
\vspace{-2mm}
\resizebox{\linewidth}{!}{
\begin{tabular}{l c c c c c}
\toprule
\textbf{Init.}
  & \textbf{SSIM} $\uparrow$
  & \textbf{DreamSim} $\downarrow$
  & \textbf{Edge F1} $\uparrow$
  & \textbf{DINO} $\uparrow$
  & \textbf{LPIPS} $\downarrow$ \\
\midrule
Random
  & 0.6718
  & 0.2459
  & 0.7975
  & 0.6378
  & 0.4034 \\
\textbf{Guided}
  & \textbf{0.7375}
  & \textbf{0.1744}
  & \textbf{0.8067}
  & \textbf{0.7328}
  & \textbf{0.3539} \\
\bottomrule
\end{tabular}
}
\end{table}

\paragraph{Layer-wise fusion.}
Tab.~\ref{tab:abl_fusion} shows the complementary role of both layers. Texture-only loses structural coherence, while structure-only preserves edges but lacks tonal richness. Fusion gives the best DINO, DreamSim, and Aesthetic scores while retaining strong Edge F1, supporting the proposed decoupled formulation.

\begin{table}[t!]
\centering
\caption{\textbf{Fusion ablation.} DINO/DreamSim use $I_{\text{tex}}$; Edge F1 uses $I_{\text{edge}}$. Best in \textbf{bold}.}
\label{tab:abl_fusion}
\vspace{-2mm}
\resizebox{\linewidth}{!}{
\setlength{\tabcolsep}{5pt}
\begin{tabular}{l c c c c}
\toprule
\textbf{Variant}
  & \textbf{DINO} $\uparrow$
  & \textbf{DreamSim} $\downarrow$
  & \textbf{Edge F1} $\uparrow$
  & \textbf{Aesthetic} $\uparrow$ \\
\midrule
Texture only
  & 0.7328
  & 0.1744
  & 0.8067
  & 5.0715 \\
Structure only
  & 0.5800
  & 0.2918
  & \textbf{0.8841}
  & 4.8718 \\
\textbf{Fusion}
  & \textbf{0.7372}
  & \textbf{0.1712}
  & 0.8629
  & \textbf{5.3507} \\
\bottomrule
\end{tabular}
}
\end{table}

\section{Conclusion}

We presented SketchForge, a training-free framework that converts raster images into editable SVG sketches through decoupled structure and texture optimization. Contour- and attention-guided initialization improves convergence and fidelity, while multiply fusion balances sharp geometry with tonal detail. Extensive experiments demonstrate that SketchForge consistently outperforms existing methods in structural fidelity, stylistic similarity, and overall sketch quality. Beyond standard vectorization, SketchForge supports color-preserving vectorization and reference-guided style transfer, offering a versatile tool for vector design and content creation. Future work will explore more efficient optimization and finer user control over individual strokes and semantic regions.

\begin{acks}
This research is funded by Vietnam National University - Ho Chi Minh City (VNU-HCM) under Grant Number B2026-18-17.
\end{acks}

\bibliographystyle{ACM-Reference-Format}
% \balance
\bibliography{references}

\end{document}